\documentclass[10pt,twocolumn,letterpaper]{article}

\usepackage[applications]{wacv}      
\usepackage[accsupp]{axessibility}
\usepackage{times}
\usepackage{epsfig}
\usepackage{xcolor} 
\usepackage{graphicx}
\usepackage{amsmath}
\usepackage{amssymb}
\usepackage{bm} 
\usepackage{caption}
\usepackage{url}            
\usepackage{amsfonts}       
\usepackage{nicefrac}       
\usepackage{microtype}      
\usepackage{comment}
\usepackage{booktabs}
\usepackage{multirow}
\usepackage{color,colortbl}
\usepackage{cuted}
\usepackage[ruled,vlined]{algorithm2e}

\usepackage[pagebackref=true,breaklinks=true,letterpaper=true,colorlinks,bookmarks=false]{hyperref}

\usepackage[capitalize]{cleveref}
\crefname{section}{Sec.}{Secs.}
\Crefname{section}{Section}{Sections}
\Crefname{table}{Table}{Tables}
\crefname{table}{Tab.}{Tabs.}

\def\wacvPaperID{2469} 
\def\confName{WACV}
\def\confYear{2024}

\newcommand{\norm}[1]{\left\lVert#1\right\rVert}

\newcommand{\eye}[0]{\mathbf{I}}
\newcommand{\be}[0]{\boldsymbol{\epsilon}}

\newcommand{\bR}[0]{\mathbb{R}}
\newcommand{\bE}[0]{\mathop{\mathbb{E}}}

\newcommand{\bz}[0]{\mathbf{z}}

\newcommand{\x}[0]{\mathbf{x}}

\newcommand{\bara}[0]{\bar{\alpha}}

\newcommand{\GNU}[0]{\mathcal{N}(\mathbf{0}, \eye)}

\newlength\savedwidth
\newcommand\whline[1]{\noalign{\global\savedwidth\arrayrulewidth
                               \global\arrayrulewidth #1} %
                      \hline
                      \noalign{\global\arrayrulewidth\savedwidth}}

\makeatletter
\renewcommand{\paragraph}{%
  \@startsection{paragraph}{4}%
  {\z@}{0.60ex \@plus 1ex \@minus .15ex}{-0.8em}%
  {\normalfont\normalsize\bfseries}%
}
\makeatother

\begin{document}

\title{Personalized Face Inpainting with Diffusion Models by Parallel Visual Attention}

\author{
Jianjin Xu$^1$ \quad Saman Motamed$^{1,3}$ \quad Praneetha Vaddamanu$^1$ \quad Chen Henry Wu$^1$ \\ 
Christian Haene$^2$ \quad Jean-Charles Bazin$^2$ \quad Fernando de la Torre$^1$ \\
$^1${\it Carnegie Mellon University} \\
$^2${\it Independent Researcher} \quad $^3${\it INSAIT, Sofia University}
}

\twocolumn[{%
\renewcommand\twocolumn[1][]{#1}%
\maketitle
\begin{center}
    \centering
    \captionsetup{type=figure}
    \vspace{-5mm}
    \includegraphics[width=0.99\textwidth]{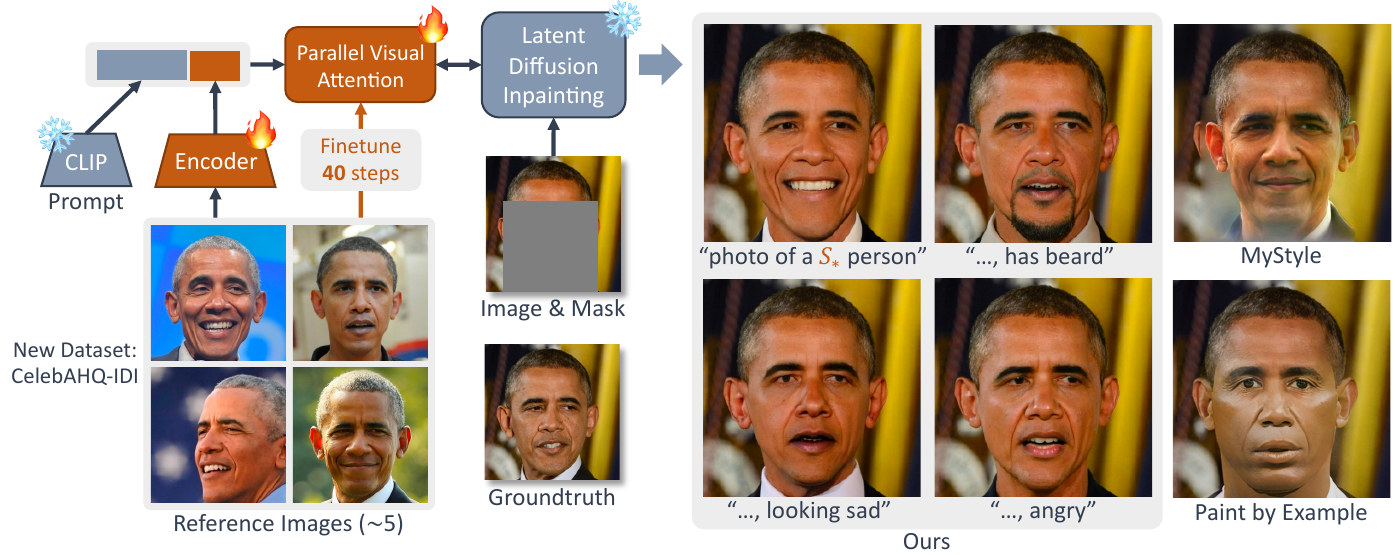}
    \vspace{-2mm}
    \captionof{figure}{
    We address the task of identity-preserving language-controllable face inpainting by adding Parallel Visual Attention (PVA) to a pretrained diffusion model. PVA enhances the diffusion model to condition on the reference images, thereby preserving the identity.
    PVA achieves the best identity similarity and image quality compared to several baselines including MyStyle~\cite{nitzan2022mystyle} and Paint by Example~\cite{yang2022paint}, even when editing the image with a prompt that changes the expression or iconic changes (e.g., beard, lipstick, changing hair style).
    }
    \label{fig:teaser}
\end{center}%
}]

\maketitle

\begin{abstract}

    Face inpainting is important in various applications, such as photo restoration, image editing, and virtual reality.
    Despite the significant advances in face generative models, ensuring that a person's unique facial identity is maintained during the inpainting process is still an elusive goal.
    Current state-of-the-art techniques, exemplified by MyStyle, necessitate resource-intensive fine-tuning and a substantial number of images for each new identity.
    Furthermore, existing methods often fall short in accommodating user-specified semantic attributes, such as beard or expression.

    To improve inpainting results, and reduce the computational complexity during inference, this paper proposes the use of Parallel Visual Attention (PVA) in conjunction with diffusion models.
    Specifically, we insert parallel attention matrices to each cross-attention module in the denoising network, which attends to features extracted from reference images by an identity encoder.
    We train the added attention modules and identity encoder on CelebAHQ-IDI, a dataset proposed for identity-preserving face inpainting.
    Experiments demonstrate that PVA attains unparalleled identity resemblance in both face inpainting and face inpainting with language guidance tasks, in comparison to various benchmarks, including MyStyle, Paint by Example, and Custom Diffusion.
    Our findings reveal that PVA ensures good identity preservation while offering effective language-controllability.
    Additionally, in contrast to Custom Diffusion, PVA requires just 40 fine-tuning steps for each new identity, which translates to a significant speed increase of over 20 times.

\end{abstract}

\section{Introduction}


The task of reconstructing absent regions in face images (i.e., face inpainting) is key to various fields, such as virtual reality, photo editing, and photo restoration.
 In recent years, as masks became a common sight due to COVID-19, many photographs captured at social gatherings or tourist attractions featured individuals wearing masks. There is a growing interest in digitally removing these masks to reveal the person's true appearance beneath. Beyond simply filling in the covered areas, there's a demand for manipulating the restored image in ways like altering facial expressions through descriptive language—a task defined as identity-preserving, language-controllable face inpainting. This technology also has broader implications, like removing sunglasses in a personal photo, inpaiting the eyes in VR meetings, and other variety of restoration or editing purposes.

Maintaining a person's recognizable features is crucial in face inpainting tasks. Take the example of individuals aiming to upload restored images on social platforms or estimate the face of a user with VR glasses;  in these applications it's essential that the user can still be recognized. Standard face inpainting tools that lack the ability to retain the user's likeness would be of minimal practical benefit. On the other hand, current strategies for identity-conserving face inpainting often rely on one or several reference photos of the individual. However, the integration of these reference photos into the inpainting process poses a significant technical challenge that has yet to be fully solved.

 A SOTA technique for identity-preserving face inpainting is MyStyle~\cite{nitzan2022mystyle}, which finetunes a pre-existing StyleGAN model using several reference photos of a particular individual. Nevertheless, MyStyle necessitates more than 40 images to maintain image quality; reducing the number of images results in lower quality outputs. In contrast, newer methods that use diffusion processes for personalizations~\cite{gal2022image,ruiz2022dreambooth}, like Custom Diffusion~\cite{kumari2022multi}, have shown remarkable success in customizing a diffusion model using as few as five images. These models excel at producing highly accurate and linguistically controllable images of specific subjects, including people. While these approaches have not yet been specifically adapted for inpainting tasks in existing research, our study has taken the initiative to apply these personalization techniques to face inpainting. We found that they deliver promising results. However, a significant drawback of these methods is the computational cost. For example, adapting MyStyle to a new individual takes several minutes, whereas diffusion-based personalization methods require upwards of four hours on an A4000 GPU.

This study presents a new approach to reduce the computational expense associated with identity-preserving and language-controllable face inpainting. We introduce an auxiliary channel alongside diffusion models, coupled with a new component, the PVA. This new workflow is illustrated in~\cref{fig:teaser}. For each cross-attention~\cite{vaswani2017attention} in the denoising UNet~\cite{ronneberger2015u}, we introduce a new set of $\{\mathbf{Q}', \mathbf{K}', \mathbf{V}'\}$ matrices that attend to visual features extracted with a vision encoder.
In training, we freeze the denoising UNet and only train the vision encoder and those new matrices.
As a result, we only need 40 steps of finetuning for a new subject in inference, which costs less than 1 minute on a single GPU, resulting in over 20 times acceleration over Custom Diffusion~\cite{kumari2022multi}.

Assessing the quality of identity-preserving face inpainting requires a dataset that offers a variety of identities, multiple reference images for each identity, and high-resolution images. Currently, there isn't a definitive benchmark for such evaluations. CelebA-IDD~\cite{dolhansky2018eye} could have been a candidate, but it falls short in resolution and is no longer accessible. To bridge this gap, we have developed a new benchmark dataset named CelebAHQ-IDI. This dataset is curated from the existing CelebAHQ dataset~\cite{lee2020maskgan}, sorted based on the availability of multiple reference images for individual identities. Additionally, we have created semantic occlusion masks that conceal parts of the face, such as the lower half or the eyes and eyebrows, to mimic real-life occlusions. This dataset will serve as the new standard for evaluating identity-preserving face inpainting tasks.

PVA was trained and tested using a subset of the CelebAHQ-IDI dataset, specifically CelebAHQ-IDI-5, which provides five reference images for each identity. We trained PVA on the training partition of CelebAHQ-IDI-5 and evaluated it on the test partition, which contains identities that PVA had not previously encountered. The effectiveness of PVA in inpainting was assessed from two critical perspectives: how well it preserved the identity and the overall quality of the image output. These were quantitatively measured using a pretrained face recognition network to determine identity preservation and Frechet Inception Distance (FID~\cite{heusel2017ttur}) and Kernel Inception Distance (KID)~\cite{binkowski2018kid} scores for image quality. PVA's performance was benchmarked against five other methods, including the notable MyStyle and Custom Diffusion models. Our results indicated that PVA surpassed all the baseline methods, achieving the highest scores in identity preservation and the lowest in FID, confirming its superior performance in both preserving identity and maintaining high image quality.

Finally, to assess the language-directed editing capabilities of our method, we crafted 15 distinct prompts aimed at altering facial expressions, actions, and accessories, among other features. We quantified the degree of alignment between the modified image and the textual prompt using the CLIP score metric~\cite{hessel2021clipscore}. Our findings reveal a balancing act between maintaining the subject's identity and achieving the desired linguistic edits. Our PVA approach achieved the best results in preserving identity. At the same time, it offered language control on par with other methods like Textual Inversion and Custom Diffusion.

\section{Related Work}

\paragraph{Generative Adversarial Networks (GANs).}
The classic architecture of GANs~\cite{goodfellow2014generative,radford2015unsupervised} consists of a generator and a discriminator.
The discriminator is trained to distinguish generated images and guides the training of the generator.
Currently, StyleGAN~\cite{karras2019style,karras2020analyzing} models hold SOTA generation quality on aligned image domains, like face, car, cat, \etc~\cite{sauer2022styleganxl,sauer2021projected}.
Besides image generation, GANs also empower various applications like image-to-image translation~\cite{Isola2017pix,zhu2017unpaired,choi2018stargan,park2019gaugan,richardson2021encoding}, image inpainting~\cite{dolhansky2018eye,zhao2018identity,lu2022diverse,nitzan2022mystyle}, and image editing~\cite{bau2018gan,abdal2019image2stylegan,zhu2020indomain,shen2020interfacegan,xu2021linear,shen2021closed,roich2022pivotal,xu2022extracting}.
The Pivotal Tuning Inversion (PTI)~\cite{roich2022pivotal} is an important technique to adapt a generic GAN to a customized object for image editing purposes.
MyStyle~\cite{nitzan2022mystyle} builds on top of PTI and finetunes a GAN on the images of a specific person.
MyStyle achieves good results in identity-preserving inpainting, super-resolution, and editing.
However, it requires 40+ images for each person and fewer images result in a severe loss of image quality.
In comparison, our method only requires 5 images.

\paragraph{Diffusion models.} Diffusion models~\cite{ho2020ddpm,song2019generative,song2020score} generate data by reversing a data corruption process. Recently, latent diffusion models~\cite{rombach2022high} have been shown to be effective for high-resolution image synthesis tasks.
Among these tasks, text-to-image generation \cite{nichol2021glide,ramesh2022dalle2} aims at generating faithful images based on a text prompt.
To adapt a generic text-to-image diffusion model to generate images of a specific object, recent work proposed to finetune text embeddings~\cite{gal2022image} or the diffusion model itself \cite{ruiz2022dreambooth,kumari2022multi}.
However, previous adaptation methods require finetuning for hundreds or thousands of steps, making it inefficient in practical applications.
In comparison, our adaptation method only needs 40 steps of finetuning for the adaptation. 

\paragraph{Identity-preserving face inpainting.}
Most existing methods for identity-preserving face inpainting use a GAN architecture.
The GAN is usually augmented with a pathway that incorporates features from an exemplar image~\cite{dolhansky2018eye,zhao2018identity,li2020ifgan,lu2022diverse}.
Dolhansky~\etal~\cite{dolhansky2018eye} proposes the ExGAN architecture for eye inpainting only.
Zhao~\etal~\cite{zhao2018identity} is trained with 128px only images.
EXE-GAN~\cite{lu2022diverse} proposes to encode an exemplar image into the style vector of a StyleGAN-like inpainting network.
However, it only shows qualitative results for identity-preserving inpainting and has no quantitative evaluation.
At the time of this work, EXE-GAN has not been open-sourced and we could not compare to it.
The closest existing works to ours are MyStyle and the diffusion-based personalization algorithms.
We use them as baselines for comparisons.

\section{Method}\label{sec:method}

\subsection{Problem Definition}\label{subsec:prob_def}

We address the task of identity-preserving language-controllable face inpainting.
Given a set of reference images ($\mathcal{R}_p$) and a set of inference images and masks ($\mathcal{I}_p$) for identity $p$, we aim to inpaint the set of masked images ($\mathcal{C}_p$) such that the inpainted images can still be perceived as identity $p$.
We define $\mathcal{R}_p=\{ \x_r \}_{i=1}^{N^\text{ref}_p}$, where $N^\text{ref}_p$ is the number of reference images for identity $p$; $\mathcal{I}_p=\{ (\x_i, \{\mathbf{m}_{i,j}\}_{j=0}^{N^\text{mask}_{p,i}})\}_{i=1}^{N^\text{infer}_p}$, where $\mathbf{m}_{i, j}$ is the $j$-th corruption mask for the $i$-th image in identity $p$; and $\mathcal{C}_p = \{\x_i \odot \mathbf{m}_{i, j}\}$.
Additionally, we consider providing language control over the inpainted content.

\subsection{Background}

\paragraph{Denoising Diffusion Probabilistic Models.}
DDPM models data as a sequence $\{\x_t\}_{t=1}^T$ where Gaussian noise is gradually added into the original data $\x_0$.
In time step $t$, Gaussian noise with variance $\beta_t$ is injected,
\begin{equation}\label{eq:ddpm_fwd_step}
    \x_{t+1} = \sqrt{1 - \beta_t} \x_t + \sqrt{\beta_t} \be,
\end{equation}
where $\be \sim \GNU$ and $0 < \beta_t < 1$.
As $t$ increases, the original data $\x_0$ gradually disappears and $\x_t$ approximates a normal distribution.
Multiple diffusion steps can be combined and expressed in a concise formulation:
\begin{equation}\label{eq:ddpm_marginal}
    \x_{t} = \sqrt{\bara_t} x_0 + \sqrt{1 - \bar{\alpha}}_t \be_t,
\end{equation}
where $\bar{\alpha}_t = \Pi_{i=1}^{t} (1 - \beta_i)$, indicating the down-weighting factor of the data term as a result of diffusion.

To generate new data points, DDPM reverses the corruption process by learning a denoising network, $\be_\theta(\x_t, t)$, that tries to predict the noise added to $\x_t$.
The network is trained to minimize the Denoising Score Matching (DSM) loss,
\begin{equation}\label{eq:dsm}
    \displaystyle
    \mathcal{L}_\text{DSM} = \bE_{\x_0, t, \be \sim \GNU} \left[ \norm{\be - \be_\theta(\x_t, t)}_2^2 \right],
\end{equation}
where $\x_t$ is perturbed from $\x_0$ under noise $\be$ and $t$ is uniformly sampled from all possible time steps.
There are plenty of sampling algorithms available for DDPM~\cite{lu2023dpmsolver,zhang2022fast} and one of the most commonly used algorithms is DDIM~\cite{song2021ddim}.
We use DDIM sampling throughout our experiments.

\begin{figure*}[t]
    \centering
    \includegraphics[width=0.8\linewidth]{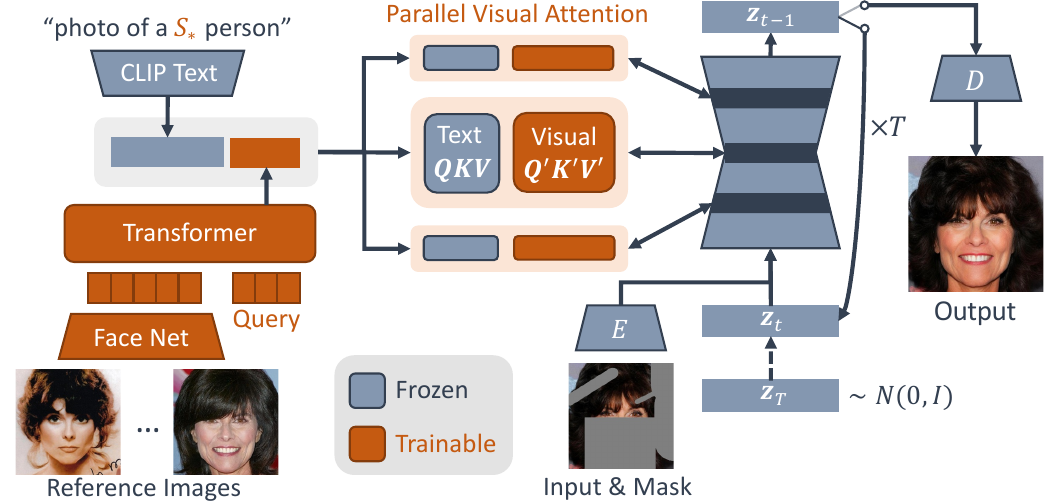}
    \vspace{-0.3cm}
    \caption{The proposed PVA pathway for incorporating reference images into a pretrained diffusion model.
    Each PVA module is modified from a cross-attention module (shown as a dark stripe on the U-Net) by adding a new set of $QKV$ matrices.
    All pretrained parameters of the denoising U-Net are frozen in training.
    }
    \label{fig:method}
\end{figure*}

\paragraph{Latent Diffusion Models.}
LDM~\cite{rombach2022high,shi2022divae} is proposed to reduce the training cost of DDPM by building diffuison models on the latent space of Variational Auto-Encoder (VAE)~\cite{kingma2014auto}.
If we denote the encoder by $\mathbf{E}_V(\cdot)$ and the decoder by $\mathbf{D}_V(\cdot)$, the encoding and decoding process are $\bz_t = \mathbf{E}_V(\x_t)$ and $\x_t = \mathbf{D}_V(\bz_t)$, respectively.
The LDM is often conditioned on text prompts through cross-attention~\cite{vaswani2017attention} modules, which attend to the text features.
The text features are extracted with a pretrained text encoder (usually CLIP~\cite{radford2021clip}), denoted by $\mathbf{y}_i = \mathbf{E}_T(T_i)$.
In the inpainting task, LDM is also conditioned on the occlusion mask and the occluded image.
The image and mask are directly concatenated to the input of LDM, denoted by $\tilde{\bz}_t = \bz_t || u_\downarrow(\mathbf{m}) || \mathbf{E}_V(\mathbf{m} \odot \x_0)$, where $u_\downarrow(\mathbf{m})$ means to downsample the mask $\mathbf{m}$ to the resolution of $\bz_0$.
The encoder for the masked image is the same as the one used for clean images.

Our method is built on top of the Latent Diffusion Inpainting (LDI) model, which is conditioned on texts, images, and masks.
LDI is initialized from a pre-trained latent diffusion checkpoint and then trained with the inpainting LDM objective:
\begin{equation}\label{eq:cdsm}
    \displaystyle
    \mathcal{L}_\text{LDM} = \bE\limits_{\substack{\bz_0, \mathbf{y}, \mathbf{m}, t, \be}} \left[ \norm{\be - \be_\theta(\tilde{\bz}_t, \mathbf{y}, t)}_2^2 \right].
\end{equation}
During training, the conditioning is dropped 10\% of the time for classifier-free guidance~\cite{ho2021classifierfree}.

\paragraph{Personalized Diffusion Models.}
There are three typical methods for the personalization of diffusion models at the time of submission.
Textual Inversion (TI)~\cite{gal2022image} finetunes the text embedding for a personalized token initialized from the original category token.
For example, the photo for a person can be described as ``A photo of $S_*$'', where the embedding of $S_*$ is the parameter to be optimized.
DreamBooth~\cite{ruiz2022dreambooth} finetunes the whole diffusion model and uses prompts with rare tokens as modifiers, \eg, ``A photo of $S_*$ person''.
Custom Diffusion~\cite{kumari2022multi} uses the same style of prompts as DreamBooth, but only finetunes the cross-attention modules and additionally tunes the embedding of the rare token.

It is observed that the diffusion models easily overfit the reference images, reducing the quality and diversity of generated images.
Therefore, both DreamBooth and Custom Diffusion need to be trained with a prior regularization loss to fight against the overfitting issue.
This means that an additional set of regularization images needs to be collected, making the algorithm more complicated.

\subsection{Parallel Visual Attention Pathway}\label{subsec:learning}

We observe two main limitations in existing methods.
First, high inference costs.
Whenever a new object is personalized, they necessitate a computationally expensive finetuning process.
Second, additional data costs due to the prior regularization loss.
We propose to reduce these costs with the Parallel Visual Attention (PVA) pathway.
The PVA pathway consists of two components.
First, a feed-forward encoder that extracts identity features from reference images.
Second, the PVA module that allows the denoising network to condition on visual features without changing existing parameters.
The pipeline is shown in \cref{fig:method}.

\subsubsection{Identity Encoder}

It is common practice to accelerate in-inference optimizations by feed-forward networks~\cite{johnson2016perceptual,huang2017arbitrary,zhu2020indomain,xu2021frame}.
A well-known example is GAN inversion~\cite{abdal2019image2stylegan,karras2020analyzing,zhu2020indomain,richardson2021encoding}, where an encoder is trained to predict the latent code given an image or segmentation.
The predicted latent code is a good initialization and thus is able to accelerate the inversion process.
Inspired by this, we also use a feed-forward model to accelerate the finetuning process in model personalization.

\paragraph{Visual feature conditioning.}
There are various ways to incorporate visual features into diffusion models.
Image-augmented diffusion models~\cite{sheynin2022knn,blattmann2022semi,chen2022reimagen} concatenate the image features to text features directly.
Paint by Example~\cite{yang2022paint} removes the text features and only keeps the visual features.
ControlNet~\cite{zhang2023adding} trains a siamese network to learn the residual to refine the original diffusion network.
However, it assumes the condition to be spatially aligned with the generated image, such as edge maps and depth maps.
In our task, the reference images might have a different pose than the image to be inpainted.
Therefore, we choose to adopt the concatenation scheme.

\paragraph{Identity feature extractor.}
Existing methods mostly use a pretrained CLIP vision encoder as the visual feature extractor~\cite{sheynin2022knn,blattmann2022semi,chen2022reimagen} due to its versatility.
However, CLIP is not trained to distinguish the nuanced differences in human faces.
So we use a pretrained face recognition network (referred to as FaceNet for convenience) as the feature extractor.

The extracted features from the FaceNet are further processed by a transformer~\cite{vaswani2017attention}.
As shown in~\cref{fig:method}, the inputs of the transformer are the FaceNet features from $M$ images and $N_\text{query}$ trainable query tokens.
The output features of query tokens are treated as visual features and later concatenated to text features.
We discard positional encoding in input because the extracted features should be invariant to the ordering of reference images.

\subsection{Parallel Visual Attention}\label{subsec:pva}

Personalizing a pretrained diffusion model on a few images of a specific object is prone to overfitting, resulting in uniform backgrounds, reduced diversity, degraded quality, \etc~\cite{ruiz2022dreambooth}.
We think the main cause of overfitting is the large capability of finetuned parameters.
Thus, we propose to finetune as fewer parameters as possible and even freeze the pretrained model.

Our solution is the Parallel Visual Attention module.
The PVA module is modified from the cross-attention module and we first introduce its formulation here.
For convenience, we use the formulation of single-head attention which can be trivially extended to the multi-head case.
A cross-attention module consists of $\mathbf{Q}, \mathbf{K}, \mathbf{V} \in \bR^{D \times D}$ matrices.
Given the flattened input feature map $\mathbf{F}_l \in \bR^{HW \times D}$ at layer $l$ and the conditional features $\mathbf{G} \in \bR^{L \times D}$, the cross-attention computes the output as
\begin{align}
    \displaystyle
    & \mathbf{F}_{l+1} = \mathbf{M} \mathbf{G} \mathbf{V}, \\
    & \mathbf{M} = \text{Softmax}\left[ \frac{\mathbf{F}_l \mathbf{Q} \cdot (\mathbf{G} \mathbf{K})^T}{\sqrt{D}} \right].
\end{align}
The PVA module introduces a new set of attention matrices, $\{\mathbf{Q}', \mathbf{K}', \mathbf{V}'\}$, which attend to the visual features and compete with the attention on text features.
PVA computes the output as follows,
\begin{align}
    & \mathbf{F}_{l+1} = \mathbf{M} \left [ \mathbf{G}_T \mathbf{V}, \mathbf{G}_V \mathbf{V}' \right ], \\
    & \mathbf{M} = \text{Softmax} ( [\mathbf{S}_T, \mathbf{S}_V ] ), \\
    & \mathbf{S}_T = \frac{\mathbf{F}_l \mathbf{Q} \cdot (\mathbf{G}_T \mathbf{K})^T}{\sqrt{D}}, \\
    & \mathbf{S}_V = \frac{\mathbf{F}_l \mathbf{Q}' \cdot (\mathbf{G}_V \mathbf{K}')^T}{\sqrt{D}},
\end{align}
where $[\cdot, \cdot]$ denotes tensor concatenation, and  $\mathbf{G}_T$, $\mathbf{G}_V$ denotes the text features and visual features, respectively.
The PVA module computes the textual and visual attention scores, $\mathbf{S}_T$ and $\mathbf{S}_V$, using two separate sets of attention matrices, and then applies softmax on the concatenated scores.
The output is obtained by weighted sum over text features and visual features transformed by $\mathbf{V}$ or $\mathbf{V}'$, separately. 

A notable characteristic of the PVA module is that when there is no visual feature, the PVA module falls back to the original cross-attention module, and the denoising network becomes identical to the pretrained one.



\subsection{Training}

We train the embedding of the special token, the PVA modules, the transformer, and the FaceNet.
The training objective augments the LDM objective (\cref{eq:cdsm}) with extra visual feature conditions,
\begin{equation}\label{eq:my_cdsm}
    \displaystyle
    \mathcal{L}_\text{LDM} = \bE\limits_{\substack{\bz_0, \mathbf{y}, \mathbf{m}, \{\x_r\}, t, \be}} \left[ \norm{\be - \be_\theta(\tilde{\bz}_t, \mathbf{y}, \mathbf{E}_I(\{\x_r\}), t)}_2^2 \right],
\end{equation}
where $\{\x_r\}$ is the set of reference images for the input image $\x_0$ and $\mathbf{E}_I$ is the identity encoder (\cref{subsec:learning}).
PVA does not need condition dropping because when the condition is dropped, the model falls back to the frozen pretrained model.

In training, we also need to provide plausible captions for images.
Previous methods~\cite{gal2022image,ruiz2022dreambooth,kumari2022multi} use templates to generate coarse descriptions of an object, \eg, ``A nice photo of ...''.
However, image inpainting is inherently ambiguous, such as recovering the expression of a person occluded by a mask.
We believe using detailed captions would be beneficial for training.
We use the captions provided by the CelebAHQ-Dialog dataset~\cite{CelebA-Dialog}, which contains detailed language descriptions of various facial attributes, including gender, age, expression, \etc.
As we do not assume to have access to detailed captions in inference, we alternatively sample from generic prompts and detailed prompts in training.

\subsection{Inference}

In inference, we can either deploy the trained model directly or run a lightweight finetuning to fit the unseen identity better.
This design choice is ablated in~\cref{subsec:ablation}.
The finetuning uses the same objective as in training, but only tunes the PVA modules.
The finetuning process consists of just 40 iterations, which takes less than 1 minute on a single GPU.

\section{Experiment}\label{sec:expr}

\begin{table}[t]
    \centering
    \begin{tabular}{cccc}
    \whline{1pt}
    \# Ref. & \# Infer. & Total Images  & \# IDs \\\hline
    1  & 23704 & 28208 & 4504 \\
    2  & 19254 & 26364 & 3555 \\
    3  & 15694 & 24328 & 2878 \\
    4  & 12799 & 22335 & 2384 \\
    5  & 10396 & 20211 & 1963 \\
    10 & 3387  & 10857 & 747  \\
    15 & 868   & 4468  & 240  \\
    20 & 120   & 1040  & 46   \\
    \whline{1pt}
    \end{tabular}
    \vspace{-0.2cm}
    \caption{Statistics of CelebAHQ-IDI dataset with different numbers of available reference images.
    ``\# Ref.'' and ``\# Infer.'' refer to the number of reference images for each identity and the number of total inference images, respectively.
    }
    \label{tab:idi_statis}
\end{table}

\begin{figure*}[t]
    \centering
    \includegraphics[width=0.99\linewidth]{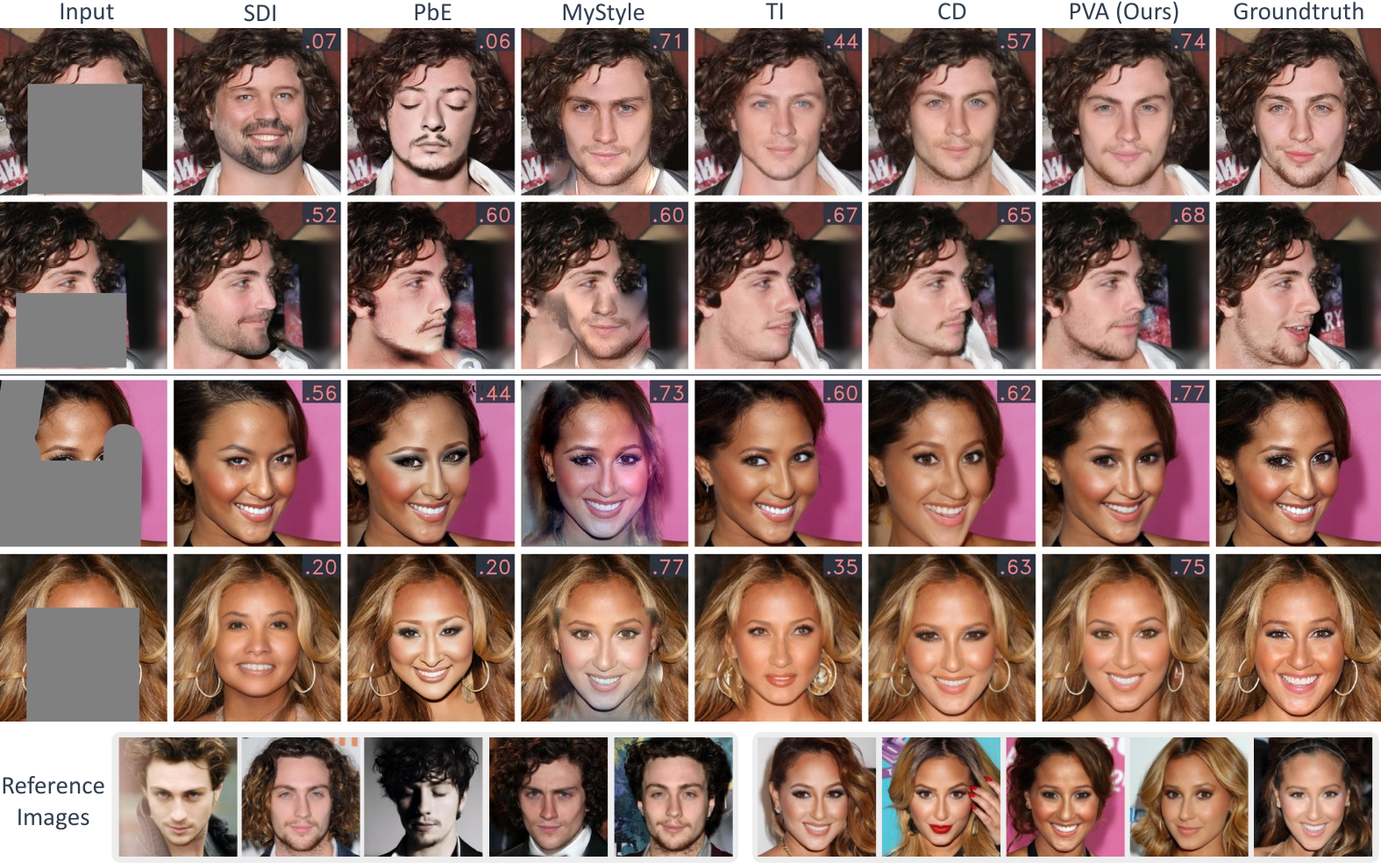}
    \vspace{-0.2cm}
    \caption{Inpainting results of PVA and baselines.
    The column tabs, ``LDI, PbE, MyStyle, TI, CD'' denotes Latent Diffusion Inpainting, Paint by Example~\cite{yang2022paint}, MyStyle~\cite{nitzan2022mystyle}, Textual Inversion~\cite{gal2022image}, and Custom Diffusion~\cite{kumari2022multi}, respectively.
    The upper right numbers of each image are the \textcolor{red}{identity similarity} ($\uparrow$) between the inpainted image and the groundtruth.
    All the methods use the same 5 reference images shown in the bottom row.
    Rows 1 and 2 use the reference images on the left and Rows 3 and 4 use those on the right.
    }
    \label{fig:qualitative_inpainting}
\end{figure*}

\subsection{Setup}\label{subsec:setup}

\paragraph{Pretrained models.}
We use a pretrained latent diffusion inpainting (LDI) model. 
For the FaceNet as a part of the identity encoder, we use the ArcFace~\cite{deng2019arcface} R50 network trained on MS1MV3 dataset~\cite{deng2019ms1mv3}.
To calculate the identity similarity, we use a different pretrained network, the CosFace~\cite{wang2018cosface} R100 network trained on the Glint360K~\cite{An2022Killing} dataset.
Both pretrained networks are obtained from InsightFace~\cite{an2020partical_fc}.


\paragraph{Dataset.}
We used CelebAHQ, CelebAHQ-Dialog~\cite{CelebA-Dialog}, and the images-of-celebs dataset\footnote{\url{https://github.com/images-of-celebs/images-of-celebs}}.
We constructed a new dataset, CelebAHQ-IDI for identity-preserving face inpainting.
CelebAHQ-IDI was built from the images in CelebAHQ and the identity annotation in CelebA~\cite{liu2015faceattributes}.
To allow a fair comparison between algorithms, we reorganized the images according to the number of reference images.
In this work, we mainly use the CelebAHQ-IDI-5 set, which has 5 reference images per identity.
We also constructed several types of semantic occlusion masks that covered ``lower face'', ``eye \& brow'', ``whole face'', and ``random'' regions.
Some examples of the masks could be found in~\cref{fig:qualitative_inpainting}.
The statistics of CelebAHQ-IDI are shown in~\cref{tab:idi_statis}.
For the detailed construction pipeline, please refer to the appendix.

\paragraph{Evaluation.}
We compared PVA to 5 baselines, the original Latent Diffusion Inpainting (LDI)~\cite{rombach2022high}, Paint by Example (PbE)~\cite{yang2022paint}, MyStyle~\cite{nitzan2022mystyle}, Textual Inversion (TI)~\cite{gal2022image} and Custom Diffusion (CD)~\cite{kumari2022multi}.
We did not compare to DreamBooth~\cite{ruiz2022dreambooth} because it needed to store the whole finetuned Diffusion model for each identity, resulting in prohibitive storage consumption.
Custom Diffusion could be regarded as an equivalence for DreamBooth as they are technically similar and have close performance~\cite{kumari2022multi}.

We evaluate the performance of face inpainting in two aspects, ID similarity and image quality.
Identity similarity was measured by the cosine similarity between the features of the inpainted images and the groundtruth images.
The features were extracted by the CosFace R100 pretrained model.
The image quality was measured by the Frechet Inception Distance (FID)~\cite{heusel2017ttur} and Kernel Inception Distance (KID)~\cite{binkowski2018kid,karras2020training} between features of inpainted images and groundtruth images.
We used the clean-fid~\cite{parmar2021cleanfid} implementation and InceptionV3~\cite{szegedy2016rethinking} as the feature extractor.

For the evaluation of language controllability, we created 15 edit prompts covering different facial expressions, makeup, action, and accessories.
We ran inpainting algorithms on images with ``whole face'' masks, ensuring a sufficient degree of freedom for editing.
The level of controllability was measured by the text alignment between the inpainted images and the target prompts.
We used the CLIP score~\cite{hessel2021clipscore} on the ViT-B/32 backbone as the metric for text alignment.
A few examples of the prompts can be found in~\cref{fig:qualitative_lang_edit}.
The full list of prompts is described in the appendix.

\paragraph{Implementation details.}
We mainly conducted training and evaluation on the CelebAHQ-IDI-5 dataset.
We trained our model for 200K iterations using the AdamW~\cite{kingma2014adam,loshchilov2017decoupled} optimizer with batch size 16, learning rate $1.6 \times 10^{-5}$, and weight decay $10^{-2}$.
In finetuning, we used the same settings and trained for just 40 steps.
We used the same sampling method across all diffusion-based methods, which was DDIM sampler with 100 steps and $\eta=0.7$.
We used PyTorch~\cite{pytorch2019} to implement the algorithms.
Other implementation details are described in the appendix. \footnote{All experiments and data processing activities are conducted at Carnegie Mellon University (CMU).}

\subsection{Results}\label{subsec:result}

We evaluated the performance of PVA on the face inpainting task and language-controlled inpainting task.
We also ablated some design choices in PVA in~\cref{subsec:ablation}.


\begin{table}
\centering
\resizebox{\linewidth}{!}{
\begin{tabular}{ccccc}
\whline{1pt}
Method            & FT. Time     & ID $\uparrow$                   & FID $\downarrow$               & \begin{tabular}[c]{@{}c@{}}KID $\downarrow$\\ {\small $\times 10^{-3}$} \end{tabular} \\ \hline
LDI              & -            & 0.359                           & 8.24                           & \textbf{2.717}                                                                       \\
Paint by Example  & -            & 0.430                           & 11.2                           & 6.089                                                                       \\ \hline
MyStyle           & $\sim$ 15min & 0.696                           & 27.7                           & 5.029                                                                         \\
Textual Inversion & $\sim$ 6h    & 0.644                           & 13.8                           & 8.404                                                                       \\
Custom Diffusion  & $\sim$ 3h    & 0.729                           & 13.9                           & 5.870                                                                           \\ \hline
PVA (Ours)        & $\sim$ 1min  & \textbf{0.741} & \textbf{8.22} &  4.289 \\                                               \whline{1pt}         
\end{tabular}}
\vspace{-0.2cm}
\caption{Quantitative comparisons of identity similarity and image quality on the CelebAHQ-IDI-5 dataset.
The finetuning costs of each method are indicated in Col ``FT. Time'', measured in single GPU (RTX A4000) time.
``ID'' stands for identity similarity.
}
\label{tab:quantitative_inpainting}
\end{table}

\subsubsection{Identity-Preserving Face Inpainting}\label{subsec:result_inpainting}

\paragraph{Qualitative results.}
The qualitative comparisons of PVA to baselines are shown in ~\cref{fig:qualitative_inpainting}.
It was observed that our method PVA outperformed all baselines.
In terms of identity similarity, all baselines had noticeable shifts in identity.
The original diffusion model, LDI, produced plausible inpainting but the inpainted person was very different since it had no information about the identity.
Paint by Example captured some characteristics of the identity like the small mustache (Row 1 Col 4) but failed to preserve the identity.
All images inpainted by MyStyle had noticeable inconsistencies, in particular the side view face photo (Row 2 Col 3).
Textual Inversion and Custom Diffusion performed relatively better among the baselines, yet they both failed to inpaint the jaw correctly, \eg, in Row 1 Col 5 and 6.
In comparison, PVA achieved the best identity similarity and image quality among all methods.

\paragraph{Quantitative results.}
The evaluation results of identity similarity and image quality are presented in~\cref{tab:quantitative_inpainting}.
It was observed that PVA outperformed all baselines on identity similarity and FID.
Although PVA was a bit below LDI on KID, the LDI had the lowest identity similarity, indicating that LDI could not preserve the identity.
The most competitive baseline was Custom Diffusion, achieving a similarity score of 0.729, yet still lower than the 0.753 score of PVA.
Moreover, Custom Diffusion also scored worse than PVA in FID and KID, showing that it has inferior image quality than PVA.
In conclusion, PVA outperformed all baselines on identity-preserving face inpainting on CelebAHQ-IDI-5.

\begin{figure}[t]
    \centering
    \includegraphics[width=0.99\linewidth, trim=0 10 0 0]{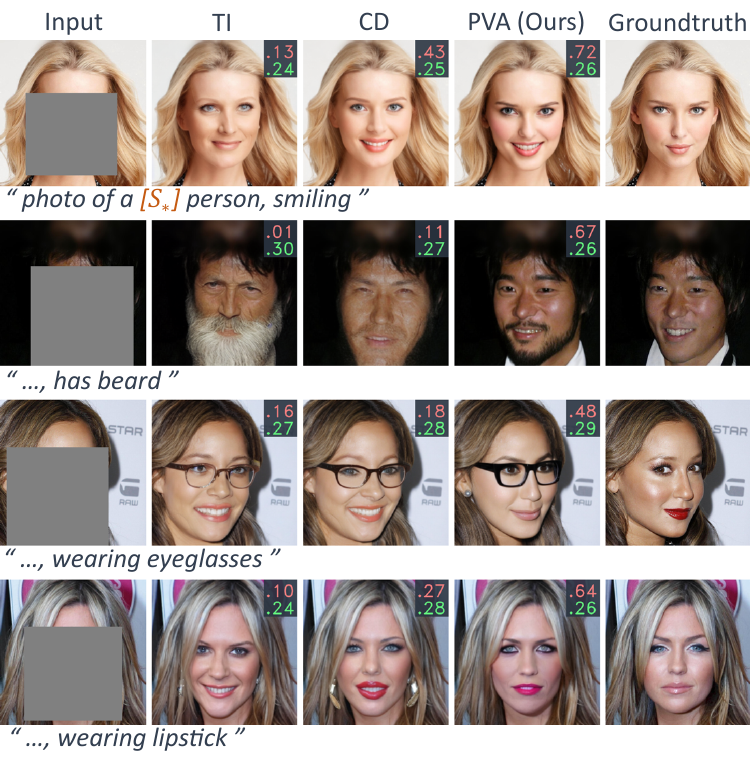}
    \caption{Qualitative comparisons of identity-preserving language-controlled inpainting.
    Prompts for editing are shown at the bottom of each row.
    We annotate the \textcolor{red}{ID similarity} (line 1)  to the groundtruth and the \textcolor{green}{CLIP similarity} (line 2) to the text prompt at the upper right of each inpainted image.
    }
    \label{fig:qualitative_lang_edit}
\end{figure}

\subsubsection{Language Controllability}\label{subsec:lang_edit}

As MyStyle and Paint by Example do not support language control, we compared PVA to Textual Inversion and Custom Diffusion on language controllability.

\paragraph{Qualitative results.}
We present language-controlled inpainting examples in~\cref{fig:qualitative_inpainting}.
Results showed that our method could control the inpainted content while preserving the identity.
In comparison, Textual Inversion and Custom Diffusion lost the target identity severely while editing the image.

\begin{figure}[t]
    \centering
    \includegraphics[width=0.78\linewidth, trim=30 30 30 30]{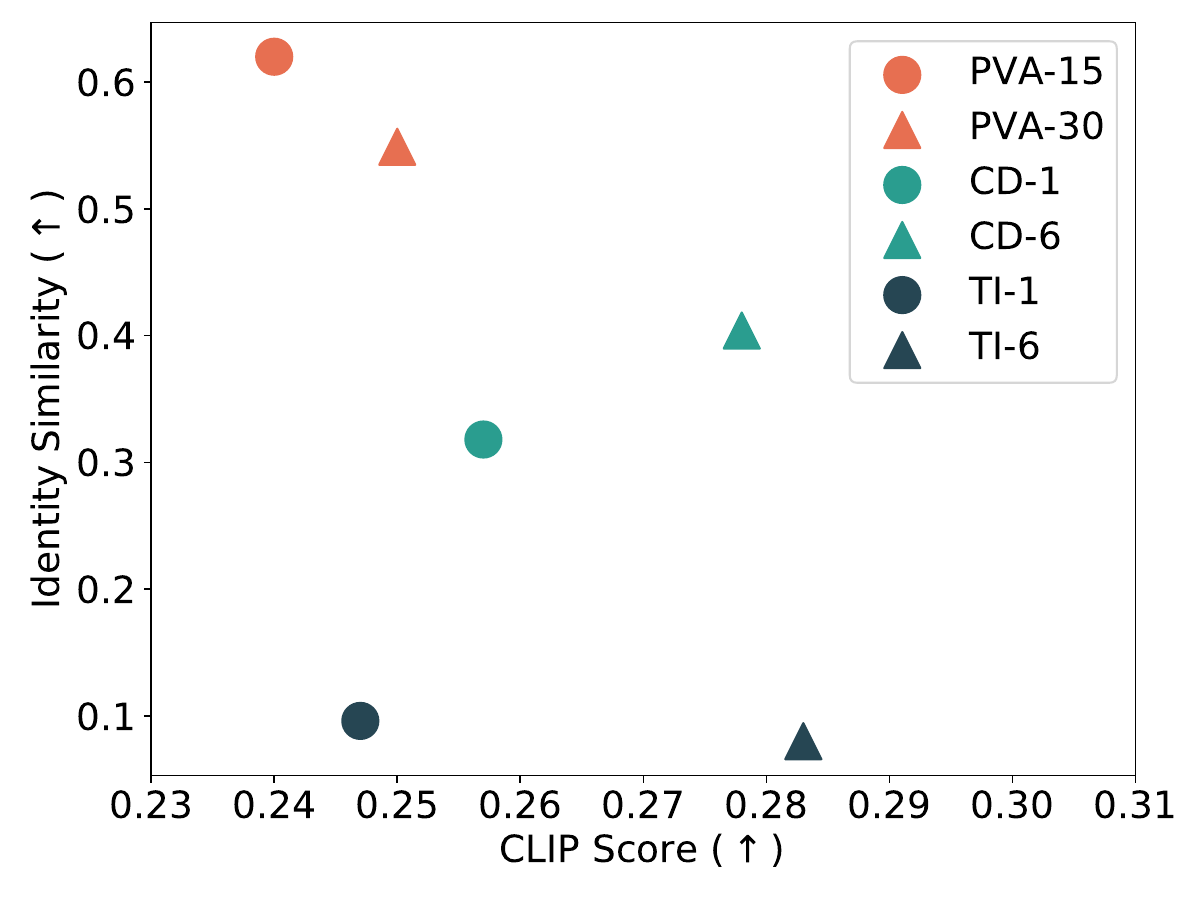}
    \caption{Identity similarity v.s. CLIP score for inpainting with language control. The suffix ``-x'' is the classifier-free guidance strength.}
    \label{fig:quantitative_lang_edit}
\end{figure}

\paragraph{Quantitative results.}
The CLIP score and identity similarity for inpainted images are shown in~\cref{fig:quantitative_lang_edit}.
We had two observations.
First, all methods had lower identity similarity compared to the inpainting-only task.
PVA, Custom Diffusion, and Textual Inversion scored around 0.6, 0.4, and 0.1, and the inpainting-only ones were 0.753, 0.729, and 0.638.
It indicated that using language control, there was a trade-off between identity similarity and prompt similarity.
Second, PVA demonstrated the best trade-off efficiency.
The identity similarity of PVA is significantly higher than the baselines.
Meanwhile, the CLIP score of PVA ranged between 0.24 and 0.25, which overlapped with CD-1 and TI-1.
The most competitive baseline, Custom Diffusion, scored around 0.4 for identity similarity.
We refer readers to the example shown in~\cref{fig:qualitative_lang_edit} Row 1 Col 3, which has a similar score of 0.43.
One could easily recognize that image as a different person than the groundtruth.
In conclusion, PVA preserves the identity significantly better while having language controllability matched to the baselines with guidance scale 1.

\subsubsection{Ablation Study}\label{subsec:ablation}

We ablated three factors that influence the identity similarity and image quality of PVA.

\begin{figure}[t]
    \centering
    \includegraphics[width=0.5\linewidth, trim=30 30 60 30]{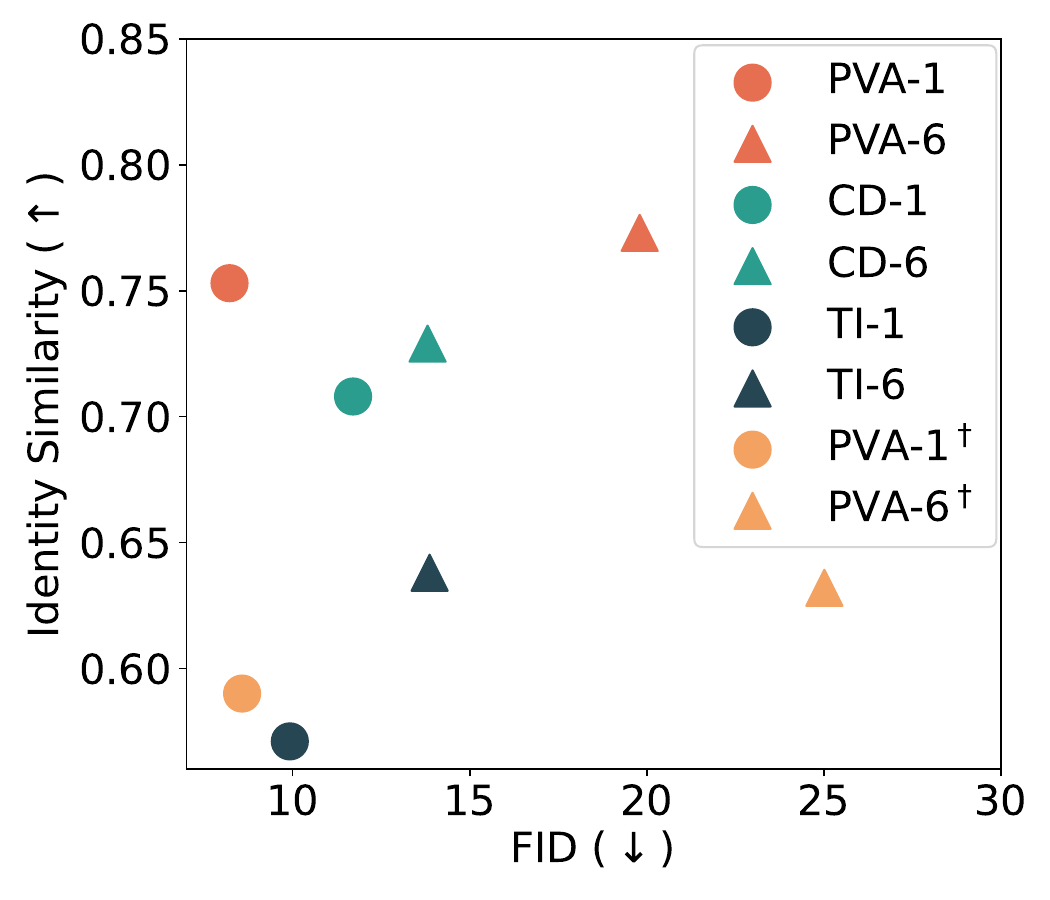}
    \caption{ID similarity and FID comparison of PVA and baselines. $\dagger$ indicates PVA without finetuning in inference.}
    \label{fig:ablation_guidance}
\end{figure}

\paragraph{1. Ablation on classifier-free guidance.}
The effect of classifier-free guidance is shown in~\cref{fig:ablation_guidance}.
We observed that a larger guidance scale increased the identity similarity, but also negatively affected the image quality.
This trade-off was present in all personalization techniques.

\paragraph{2. Ablation on finetuning.}
See the comparison between PVA and $\text{PVA}^\dagger$ (without finetuning) in~\cref{fig:ablation_guidance}.
It was observed that PVA without finetuning performed closely to Textual Inversion.
With only 40 steps of finetuning, the identity similarity of PVA was significantly improved, outperforming the baselines that need over 1K optimization steps.
Therefore, it was supported that the feed-forward component of PVA learned a good initialization for finetuning.

\begin{figure}[t]
    \centering
    \includegraphics[width=0.5\linewidth, trim=30 30 60 30]{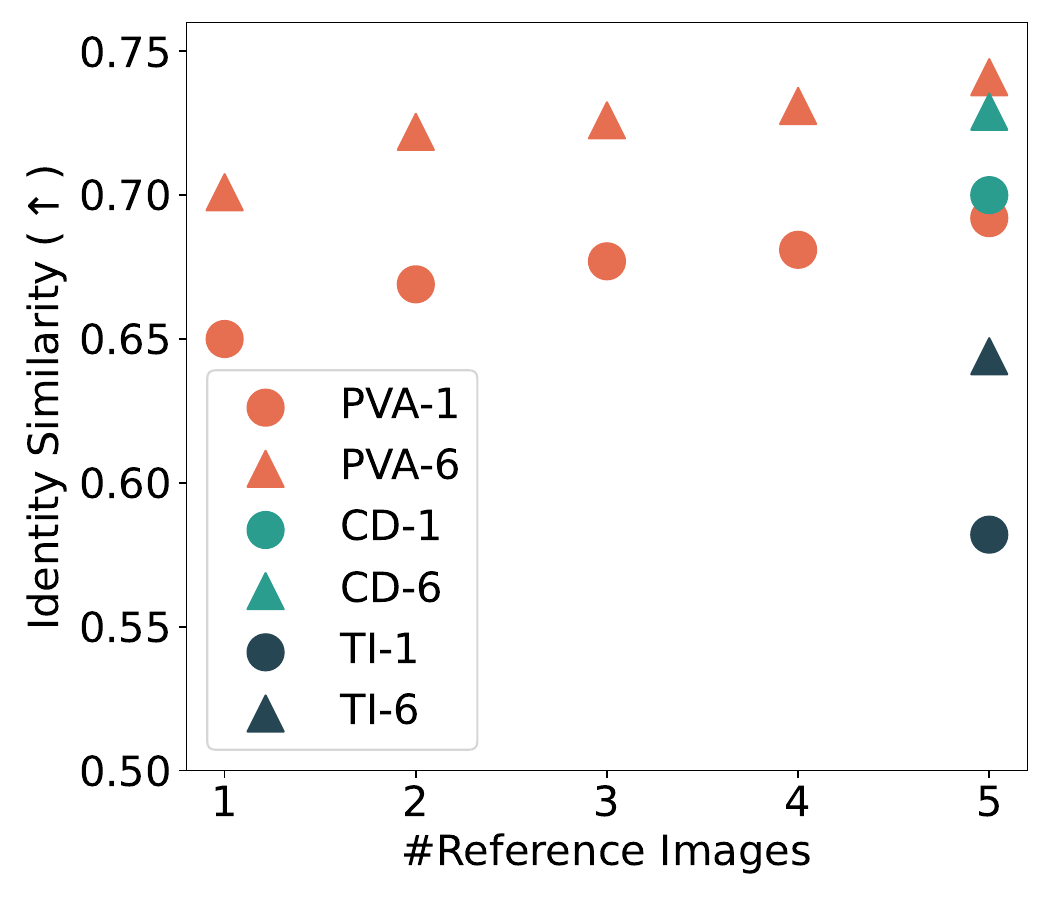}
    \caption{Ablation study on the number of reference images.}
    \label{fig:ablation_ref_num}
\end{figure}

\paragraph{3. Ablation on the number of reference images.}
We trained PVA with different numbers of reference images and presented the results in~\cref{fig:ablation_ref_num}.
As shown in the figure, more reference images resulted in better identity similarity.
Moreover, just one reference image already achieved a good result.
We speculate that the pretrained diffusion model has learned to disentangle the person's identity from other attributes.

\section{Limitation}\label{sec:limitation}
We observe that the language controllability of PVA is sacrificed to a certain degree for better identity similarity.
Though maintaining the identity similarity is indeed the top priority of this work, how to preserve the language control capability better remains an open question.

\section{Conclusion}\label{sec:conclusion}

We address the problem of identity-preserving and language-controllable face inpainting.
Our solution is the PVA pathway for diffusion models, which consists of an identity encoder and PVA modules.
We also propose a new dataset, CelebAHQ-IDI, for benchmarking identity-preserving face inpainting.
Results show that our method achieves the best identity similarity and image quality in face inpainting while being significantly faster than baselines.
In terms of language-controllability, PVA achieved a similar degree of controllability while preserving the identity better.

{\small
\bibliographystyle{ieee_fullname}
\bibliography{main}
}


\end{document}


\title{Supplementary Material for \\ Personalized Face Inpainting with Diffusion Models by Parallel Visual Attention}

\author{Jianjin Xu\\
Carnegie Mellon University\\
Pittsburgh, PA\\
{\tt\small jianjinx@andrew.cmu.edu}
\and
Second Author\\
Institution2\\
First line of institution2 address\\
{\tt\small secondauthor@i2.org}
}

\renewcommand{\figurename}{Supplementary Figure}
\renewcommand{\tablename}{Supplementary Table}

\setcounter{page}{1}

\begin{strip}
    \vspace{-13mm}
\begin{center}
 \Large
 \textbf{Supplementary Material for} \\
 \smallskip
 \smallskip
 \textbf{Personalized Face Inpainting with Diffusion Models by Parallel Visual Attention}
 \smallskip
\end{center}
\end{strip}

\section{Additional Implementation Details}

\subsection{Training of PVA Pathway}

We trained the PVA modules, the identity encoder, and the embedding of the special token.
The attention matrices of the PVA module were first initialized from the text attention matrices $\{\mathbf{Q}, \mathbf{K}, \mathbf{V}\}$ and then trained.
We used $1.6 \times 10^{-5}$ learning rate for the PVA modules and the identity encoder, but $10^{-3}$ learning rate for the special token.
Notice that we used the same token across all identities.
If we use a different token for each identity, then the special token for an unseen identity will still be unknown in inference.

We used a two-stage training strategy.
First, we trained the PVA modules, the transformer, and the special token while keeping the FaceNet frozen.
Second, we trained all these components together.
The two stages took 100K iterations each.
The reason for doing this was that the transformer and PVA were not trained at first but FaceNet was a pretrained model.
Allowing the early-stage gradients to flow into FaceNet might ruin the pretrained model.

To encourage PVA to incorporate a flexible number of reference images, we randomly sampled a subset of reference images in each training step.
Specifically, we uniformly sampled from $\{1, 2, 3, 4, 5\}$ as the number of reference images, and randomly sampled from the 5 reference images.
Also, with 0.5 probability, we replicate one of the reference images using its horizontally flipped version.
This was to align with the settings of finetuning.
See~\cref{subsec:supp_finetune} for explanations.

All images were augmented with only random horizontal flips.
We used random masks from the CelebAHQ-IDI dataset in training.
In each batch, we merged the random masks with randomly selected semantic rectangular masks.

The training was parallelized on 4 RTX A4500 GPUs and took around 70 hours.

\subsection{Finetuning}\label{subsec:supp_finetune}

Given $N_\text{ref}$ images of identity $p$, $\mathcal{R}_p=\{ \x_r \}_{i=1}^{N^\text{ref}_p}$, we finetuned the PVA and cross attention modules so that the diffusion models could adapt to this identity better.
Notice that in finetuning, we needed to use one image for inpainting and other images as reference, resulting in only $N^\text{ref}_p - 1$ reference images.
In inference, we still used the $N^\text{ref}_p$ images as reference, which would be an inconsistency between training and inference.
Therefore, we chose to ``pad'' the length of reference images to $N^\text{ref}_p$ by replicating an image.
Specifically, we randomly select one image from the $N^\text{ref}_p - 1$ references and reflected it horizontally.
When $N^\text{ref}_p=1$, however, we did not have any reference image.
In this case, we use the reflected version of the inference image as the reference.
As the finetuning was only 40 steps, we found that this did not lead to the trivial solution of copy-pasting the reference directly to the inpainted image.

We used a stratified sampling approach to reduce the variance of the gradients.
Specifically, we replicated the same batch of images $m$ times, each with a different time, $\{t_i\}_{i=1}^{m}$.
The conventional sampling approach is to sample each $t_i$ from the same uniform distribution $\mathcal{U}[0, 1]$.
In stratified sampling, we sample $t_i$ from $\mathcal{U}[\frac{i - 1}{m}, \frac{i}{m}]$.
The diffusion models at different time steps will have drastically different behavior, \eg, imagining new structures when $t$ is large and refining the details when $t$ is small.
The stratified sampling technique ensured that different time steps could be covered evenly with a small number of batch sizes, \eg, 4.

\begin{table*}[t]
    \centering
    \begin{tabular}{cc|c|c}
    \whline{1pt}
   Task & Methods & Positive Condition & Negative Condition \\\hline
\multirow{2}{*}{Inpainting-Only}   & PVA        & Photo of a person \& $\mathbf{E}_I(\{\x_r\})$ & Photo of a person \\
   & Default    & Photo of a person & $\varnothing$ \\\hline
\multirow{2}{*}{Controlled-Inpainting}   & PVA        & Photo of a person, smiling \& $\mathbf{E}_I(\{\x_r\})$ & Photo of a person \& $\mathbf{E}_I(\{\x_r\})$ \\
   & Default    & Photo of a person, smiling & $\varnothing$ \\
    \whline{1pt}
    \end{tabular}
    \caption{Comparisons between PVA and the default method on the conditions used in classifier-free guidance.}
    \label{tab:supp_classifier_free}
\end{table*}

\subsection{Reproduction of Baselines}

We used the pretrained model of LDI and Paint by Example as-is.
The Paint by Example model was obtained from their official release\footnote{\url{https://github.com/Fantasy-Studio/Paint-by-Example}}.

Textual Inversion, Custom Diffusion, and MyStyle were fine-tuned on each identity of the test set separately.
Both Textual Inversion and Custom Diffusion used a batch size of 8 and AdamW with $10^{-2}$ weight decay.
Textual Inversion was trained for 5K iterations using an effective learning rate of $10^{-2}$.
For Custom Diffusion, we trained the cross-attention modules with $8 \times 10^{-6}$ learning rate for 1K iterations.
MyStyle was trained for 1K iterations using Adam~\cite{kingma2014adam} with a learning rate of $3 \times 10^{-3}$.
In inference, MyStyle projected the image to be inpainted onto the latent space of the finetuned model, which took around 1 minute for each image.

Textual Inversion and Custom Diffusion were trained on 4 RTX A4000 GPUs, which took around 1.2 hours and 1 hour for each identity.
MyStyle was trained on a single RTX A4000 GPU and took around 15 minutes.

\subsection{Inference}

We used a slightly different setting in the classifier-free guidance.
The conventional setting of classifier-free guidance used ``photo of a person'' as the positive condition and $\varnothing$ as the negative condition.
We also used this setting in Textual Inversion and Custom Diffusion.

However, the PVA was different in that the condition had extra visual features, and the identity-related information was mostly contained in the visual component.
In light of this, we could keep the text features the same and contrast the visual features.
In the inpainting-only task, we used the ``photo of a person'' with visual features as the positive condition and used the prompt without the visual features as negative ones.
In the language-controlled inpainting task, we used the controlling prompt with visual features as the positive condition and the neutral prompt with visual features as the negative condition.
The differences are summarized in Supplementary Table~\ref{tab:supp_classifier_free}.

\subsection{Evaluation}

We evaluated the inpainting performance on four different types of semantic regions, lower face, eye \& brow, whole face, and random.
As different semantic regions might have different characteristics, we calculated the metrics for every region separately and averaged the results across all four regions.
The results per region are described in~\cref{subsec:results_per_region}.

The full list of prompts used in the language-based controlling experiment is listed in Supplementary Table~\ref{tab:full_prompt}.

\begin{table}[t]
\centering
\resizebox{\linewidth}{!}{
\begin{tabular}{lll}
\whline{1pt}
Type & Name & Prompt \\
\hline
\multirow{6}{*}{Expression} & Laughing & Photo of a person, laughing \\
& Serious & Photo of a person, serious \\
& Smile & Photo of a person, smiling \\
& Sad & Photo of a person, looking sad \\
& Angry & Photo of a person, angry \\
& Surprised & Photo of a person, surprised \\\hline
\multirow{3}{*}{Makeup} & Makeup & Photo of a person, with heavy makeup \\
& Beard & Photo of a person, has beard \\
& Lipstick & Photo of a person, wearing lipstick \\\hline
\multirow{4}{*}{Action} & Funny & Photo of a person, making a funny face \\
& Tongue & Photo of a person, putting the tongue out \\
& Singing & Photo of a person, singing with a microphone \\
& Cigarette & Photo of a person, smoking, has a cigarette \\\hline
\multirow{2}{*}{Accessory} & Eyeglass & Photo of a person, wearing eyeglasses \\
& Sunglasses & Photo of a person, wearing sunglasses \\
\whline{1pt}
\end{tabular}}
\caption{The full list of prompts used in the language-controllable inpainting experiment.}
\label{tab:full_prompt}
\end{table}

\section{Construction Pipeline of CelebAHQ-IDI}

\paragraph{Preprocessing.}
We first checked for duplicate images, including horizontally reflected duplicates.
We filtered out 403 duplicate images in total, which consisted of 199 pairs and 5 triplets.
Then we detected the facial landmarks using dlib~\cite{dlib09}. 
Two images that failed in detection were also discarded.

\paragraph{Mask generation.}
We constructed rectangular masks that covered several semantic regions of the face, including ``eye and brow'', ``lower face'', ``whole face'', \etc.
Each mask was the bounding box of the landmarks of the corresponding semantic region and was diluted 20\% in both width and height.
We also generated random masks following the protocol of LaMa~\cite{suvorov2022lama} and merged them with rectangular masks.
Specifically, we used the ``configs/data\_gen/random\_thick\_512.yaml'' configurations in the LaMa~\cite{suvorov2022lama} code base\footnote{\url{https://github.com/advimman/lama}} for generating the random masks.
We sampled 30K masks and stored them and directly sampled from these 30K masks as random masks in training.

\paragraph{Dataset split.}
We filtered all identities with images less than or equal to the reference number.
For each remaining identity, we randomly chose reference images and left the rest as inference images.
Finally, we randomly split the dataset into training, validation, and testing with ratios of 0.6, 0.1, and 0.3 based on identities.

\section{Per Region Evaluation Results}\label{subsec:results_per_region}

The identity similarity, FID, and KID per region for all methods are presented in Supplementary Tables~\ref{tab:id_per_region}, ~\ref{tab:fid_per_region}, and~\ref{tab:kid_per_region}.
We used the ``Mean'' results in the paper.
We observed that the eye \& brow region is the easiest region for inpainting and the whole face region is the hardest region.

\begin{table*}[b]
\centering
\begin{tabular}{c|cccc|c}
\whline{1pt}
Method & Lower Face & Eye \& Brow & Whole Face & Random & Mean \\\hline
LDI & 0.444 $\pm$ 0.105 & 0.613 $\pm$ 0.086 & 0.094 $\pm$ 0.088 & 0.283 $\pm$ 0.210 & 0.359 \\
PbE & 0.500 $\pm$ 0.098 & 0.638 $\pm$ 0.084 & 0.217 $\pm$ 0.100 & 0.363 $\pm$ 0.180 & 0.430 \\
MyStyle & 0.696 $\pm$ 0.134 & 0.786 $\pm$ 0.100 & 0.639 $\pm$ 0.133 & 0.661 $\pm$ 0.146 & 0.696 \\
\hline
TI-1 & 0.639 $\pm$ 0.091 & 0.759 $\pm$ 0.066 & 0.401 $\pm$ 0.132 & 0.528 $\pm$ 0.176 & 0.582 \\
TI-6 & 0.686 $\pm$ 0.098 & 0.789 $\pm$ 0.072 & 0.504 $\pm$ 0.153 & 0.597 $\pm$ 0.171 & 0.644 \\
CD-1 & 0.735 $\pm$ 0.089 & 0.823 $\pm$ 0.062 & 0.580 $\pm$ 0.131 & 0.659 $\pm$ 0.146 & 0.700 \\
CD-6 & 0.757 $\pm$ 0.094 & 0.832 $\pm$ 0.068 & 0.635 $\pm$ 0.136 & 0.694 $\pm$ 0.141 & 0.729 \\
\hline
PVA-1 & 0.634 $\pm$ 0.087 & 0.776 $\pm$ 0.064 & 0.418 $\pm$ 0.115 & 0.532 $\pm$ 0.165 & 0.590 \\
PVA-2 & 0.670 $\pm$ 0.084 & 0.797 $\pm$ 0.063 & 0.494 $\pm$ 0.114 & 0.585 $\pm$ 0.150 & 0.637 \\
PVA-4 & 0.671 $\pm$ 0.084 & 0.796 $\pm$ 0.064 & 0.505 $\pm$ 0.112 & 0.587 $\pm$ 0.144 & 0.640 \\
PVA-6 & 0.657 $\pm$ 0.091 & 0.788 $\pm$ 0.074 & 0.496 $\pm$ 0.116 & 0.586 $\pm$ 0.144 & 0.632 \\
\hline
PVA-FT-1 & 0.772 $\pm$ 0.082 & 0.856 $\pm$ 0.057 & 0.668 $\pm$ 0.116 & 0.716 $\pm$ 0.122 & 0.753 \\
PVA-FT-6 & 0.789 $\pm$ 0.096 & 0.858 $\pm$ 0.064 & 0.707 $\pm$ 0.127 & 0.740 $\pm$ 0.120 & 0.773 \\
\whline{1pt}
\end{tabular}
\caption{Comparisons of identity similarity per masked region on CelebAHQ-IDI-5 dataset.
Numbers after ``$\pm$'' indicate the standard deviation.
The ``-1'' and ``-6'' denote the classifier-free guidance strength.
``FT'' denotes finetuning on each identity for 40 iterations.
}
\label{tab:id_per_region}
\end{table*}

\begin{table*}[b]
\centering
\begin{tabular}{c|cccc|c}
\whline{1pt}
Method & Lower Face & Eye \& Brow & Whole Face & Random & Mean  \\\hline
LDI2 & 7.039 & 4.244 & 12.301 & 9.383 & 8.242 \\
PbE & 10.092 & 5.866 & 15.081 & 13.682 & 11.180 \\
MyStyle & 29.221 & 8.993 & 34.754 & 37.755 & 27.681 \\\hline
CD-1 & 6.041 & 3.709 & 7.262 & 7.288 & 6.075 \\
CD-6 & 9.540 & 4.569 & 12.969 & 11.829 & 9.727 \\
TI-1 & 6.770 & 3.746 & 9.335 & 8.821 & 7.168 \\
TI-6 & 19.370 & 5.403 & 29.811 & 23.759 & 19.586\\
\hline
PVA-1 & 8.613 & 6.296 & 9.615 & 9.766 & 8.572 \\
PVA-2 & 10.501 & 7.179 & 11.923 & 11.792 & 10.349 \\
PVA-4 & 10.289 & 6.968 & 12.129 & 12.243 & 10.407 \\
PVA-6 & 25.377 & 14.185 & 30.156 & 30.342 & 25.015 \\\hline
PVA-1 & 8.323 & 6.061 & 9.243 & 9.240 & 8.217 \\
PVA-6 & 20.18 & 12.3 &  23.2 & 23.65 & 19.8 \\
\whline{1pt}
\end{tabular}
\caption{FID per masked region for all methods on CelebAHQ-IDI-5 dataset.
For the notations in the table, please refer to the Supplementary Table~\ref{tab:id_per_region}.}
\label{tab:fid_per_region}
\end{table*}

\begin{table*}[h]
\centering
\begin{tabular}{c|cccc|c}
\whline{1pt}
Method & Lower Face & Eye & Whole Face & Random & Mean  \\\hline
LDI & 1.782 & 1.793 & 4.486 & 2.808 & 2.717 \\
PbE & 4.750 & 2.901 & 9.220 & 7.486 & 6.089 \\
MyStyle & 4.620 & 1.169 & 6.351 & 7.977 & 5.029 \\\hline
CD-1 & 5.188 & 5.629 & 5.299 & 5.637 & 5.438 \\
CD-6 & 5.279 & 5.454 & 6.120 & 6.628 & 5.870 \\
TI-1 & 5.071 & 4.775 & 5.587 & 5.151 & 5.146 \\
TI-6 & 6.735 & 5.399 & 11.487 & 9.993 & 8.404 \\\hline
PVA-1 & 4.433 & 4.090 & 4.314 & 4.637 & 4.369 \\
PVA-2 & 5.415 & 4.784 & 5.873 & 6.224 & 5.574 \\
PVA-4 & 5.729 & 4.883 & 6.737 & 7.273 & 6.155 \\
PVA-6 & 8.551 & 7.122 & 10.295 & 10.634 & 9.151 \\\hline
PVA-FT40-1 & 4.359 & 4.008 & 4.368 & 4.421 & 4.289 \\
PVA-FT40-6 & 4.600 & 4.929 & 5.282 & 4.874 & 4.921 \\
\whline{1pt}
\end{tabular}
\caption{KID ($\times 10^{-3}$) per masked region for all methods on CelebAHQ-IDI-5 dataset.
For the notations in the table, please refer to the Supplementary Table~\ref{tab:id_per_region}.
}
\label{tab:kid_per_region}
\end{table*}

{\small
\bibliographystyle{ieee_fullname}
\bibliography{egbib}
}